\documentclass[11pt,a4paper]{article}
\usepackage[hyperref]{emnlp-ijcnlp-2019}
\usepackage[utf8]{inputenc}
\usepackage[T1]{fontenc}
\usepackage[vietnamese,english]{babel}

\usepackage{times}
\usepackage{microtype}
\usepackage{amsmath}
\usepackage{mathtools}
\usepackage{microtype}
\usepackage{amsfonts}
\usepackage{amssymb}
\usepackage{tipa}
\usepackage{graphicx}

\usepackage{cleveref}
\crefname{section}{\S}{\S\S}
\Crefname{section}{\S}{\S\S}
\crefname{table}{Table}{}
\crefname{figure}{Fig.}{}
\crefname{algorithm}{Alg.}{}
\crefname{equation}{eq.}{}
\crefname{appendix}{App.}{}
\crefformat{section}{\S#2#1#3} 

\newcommand{\xx}{\mathbf{x}}
\newcommand{\ww}{\mathbf{w}}
\newcommand{\hh}{\mathbf{h}}
\newcommand{\cc}{\mathbf{c}}
\newcommand{\calL}{{\cal L}}

\newcommand{\calDtrain}{{\cal D}}
\newcommand{\bH}{\mathbf{H}}
\newcommand{\E}{\mathbb{E}}
\newcommand{\vh}{\mathbf{h}}
\newcommand{\train}{\mathcal{T}}
\newcommand{\eval}{\mathcal{E}}

\DeclarePairedDelimiter{\nint}\lfloor\rceil

\newcommand{\appropto}{\mathrel{\vcenter{
  \offinterlineskip\halign{\hfil$##$\cr
    \propto\cr\noalign{\kern2pt}\sim\cr\noalign{\kern-2pt}}}}}

\usepackage{todonotes}
\makeatletter
\newcommand*\iftodonotes{\if@todonotes@disabled\expandafter\@secondoftwo\else\expandafter\@firstoftwo\fi}  
\makeatother

\usepackage{booktabs}
\usepackage{tabularx}
\usepackage{tabulary}
\usepackage{colortbl}

\usepackage{url}
\usepackage{graphicx}
\usepackage[lofdepth,lotdepth]{subfig}
\usepackage{fancyhdr}
\usepackage{setspace}

\aclfinalcopy 

\title{
Towards Zero-shot Language Modeling}

\author{\bf Edoardo M. Ponti$^1$, Ivan Vuli\'{c}$^{1}$, Ryan Cotterell$^2$, Roi Reichart$^3$, Anna Korhonen$^1$ \\
$^1$Language Technology Lab, TAL, University of Cambridge \\
$^2$Computer Laboratory, University of Cambridge \\
$^2${Faculty of Industrial Engineering and Management, Technion, IIT} \\
$^{1,2}$\texttt {\{ep490,iv250,rdc42,alk23\}@cam.ac.uk} \\
$^3$\texttt {roiri@ie.technion.ac.il}
}

\date{}

\begin{document}
\maketitle

\begin{abstract}

Can we construct a neural model that is inductively biased towards learning {human} languages? Motivated by this question, we aim at constructing an informative prior over neural weights, in order to adapt quickly to held-out languages in the task of character-level language modeling. We infer this distribution from a sample of typologically diverse training languages via Laplace approximation. The use of such a prior outperforms baseline models with an uninformative prior (so-called `fine-tuning') in both zero-shot and few-shot settings. This shows that the prior is imbued with universal phonological knowledge. Moreover, we harness additional language-specific side information as distant supervision for held-out languages. Specifically, we condition language models on features from typological databases, by concatenating them to hidden states or generating weights with hyper-networks. These features appear beneficial in the few-shot setting, but not in the zero-shot setting. Since the paucity of digital texts affects the majority of the world's languages, we hope that these findings will help broaden the scope of applications for language technology.
\end{abstract}

\section{Introduction}
\label{sec:introduction}
With the success of recurrent neural networks and other black-box models on core NLP tasks, such as language modeling, researchers have turned their attention to the study of the inductive bias such neural models exhibit \cite{linzen-etal-2016-assessing,D18-1151,ravfogel2018can}. A number of natural questions have been asked. For example, do recurrent neural language models learn syntax \cite{D18-1151}? Do they map onto grammaticality judgments \citep{warstadt2018neural}? 
However, as \newcite{ravfogel2019studying} note, ``[m]ost of the work so far has focused on English.'' Moreover, these studies have almost always focused on training scenarios where a large number of in-language sentences are available. 

In this work, we aim to find a prior distribution over network parameters that generalize well to new human languages. 
The recent vein of research on the inductive biases of neural nets implicitly assumes a uniform (unnormalizable) prior over the space of neural network parameters \citep[\textit{inter alia}]{ravfogel2019studying}. 
In contrast, we take a Bayesian-updating approach: First, we approximate the posterior distribution over the network parameters using the Laplace method \citep{azevedo1994laplace}, given the data from a sample of \textit{seen} training languages. Afterward, this distribution serves as a prior for maximum-a-posteriori (MAP) estimation of network parameters for the held-out unseen languages.

The search for a universal prior for linguistic knowledge is motivated by the notion of Universal Grammar (UG), originally proposed by \citet{noam1959review}. The presence of innate biological properties of the brain that constrain possible human languages was posited to explain why children learn languages so quickly despite the poverty of the stimulus \citep{chomsky1978naturalistic,legate2002empirical}. In turn, UG has been connected with \citet{greenberg1963some}'s typological universals by \citet{graffi1980universali} and \citet{gilligan1989cross}: this way, the patterns observed in cross-lingual variation could be explained by an innate set of parameters wired into a language-specific configuration during the early phases of language acquisition.

Our study explores the task of character-level language modeling. Specifically, we choose an open-vocabulary setup, where no token is treated as unknown, to allow for a fair comparison among the performances of different models across different languages \cite{gerz2018language,gerz2018relation,cotterell-etal-2018-languages,Mielke:2019acl}. We run experiments under several regimes of data scarcity for the held-out languages (zero-shot, few-shot, and joint multilingual learning) over a sample of 77 typologically diverse languages.

As an orthogonal contribution, we also note that realistically we are not completely in the dark about held-out languages, as coarse-grained grammatical features are documented for most world's languages and available in typological databases such as URIEL \citep{littell2017uriel}. Hence, we also explore a regime where we condition the universal prior on typological side information.
In particular, we consider concatenating typological features to hidden states \citep{ostling2017continuous} and generating the network parameters with hyper-networks receiving typological features as inputs \citep{platanios2018contextual}.

Empirically, given the results of our study, we offer two findings. The first is that neural recurrent models with a universal prior significantly outperform baselines with uninformative priors both in zero-shot and few-shot training settings. Secondly, conditioning on typological features further reduces bits per character in the few-shot setting, but we report negative results for the zero-shot setting, possibly due to some inherent limitations of typological databases \citep{ponti2018modeling}.

The study of low-resource language modeling also has a practical impact. According to \citet{simons2017ethnologue}, 45.71\% of the world's languages do not have written texts available. The situation is even more dire for their \textit{digital} footprint. As of March 2015, just 40 out of the 188 languages documented on the Internet accounted for 99.99\% of the web pages.\footnote{ \url{https://w3techs.com/technologies/overview/content_language/all}} And as of April 2019, Wikipedia is translated only in 304 out of the 7097 existing languages. What is more, \citet{kornai2013digital} prognosticates that the digital divide will act as a catalyst for the extinction of many of the world's languages. The transfer of language technology may help reverse this course and give space to unrepresented communities.

\section{LSTM Language Models}
\label{sec:LSTM}
In this work, we address the task of \textit{character-level} language modeling. Whereas word lexicalization is mostly arbitrary across languages, phonemes allow for transferring universal constraints on phonotactics\footnote{E.g. with few exceptions \citep[sec.\ 2.2.2]{evans2009myth}, the basic syllabic structure is vowel--consonant.} and language-specific sequences that may be shared across languages, such as borrowings and cognates \citep{brown2008automated}. Since languages are mostly recorded in text rather than phonemic symbols (IPA), however, we focus on characters as a loose approximation of phonemes.

Let $\Sigma_\ell$ be the set of characters for language $\ell$. Moreover, consider a collection of languages $\mathcal{T} \sqcup \mathcal{E}$ partitioned into two disjoint sets of observed (training) languages $\mathcal{T}$ and held-out (evaluation) languages $\mathcal{E}$. Then, let $\Sigma = \cup_{\ell \in (\mathcal{\mathcal{T} \sqcup \mathcal{E})}} \Sigma_\ell$ be the union of character sets in all languages. A universal, character-level language model is a probability distribution over $\Sigma^*$.\footnote{Note that $\Sigma$ is also augmented with punctuation and white space, and distinguished beginning-of-sequence and end-of-sequence symbols, respectively.} Let $\xx \in \Sigma^*$ be a sequence of characters. We write:
\begin{equation} \label{eq:nextcharprob}
p(\xx \mid \ww) = \prod_{t=1}^n p(x_t \mid \xx_{< t}, \ww)
\end{equation}
where \textit{t} is a time step, $x_0$ is a distinguished begin\-ning-of-sentence symbol, $\ww$ are the parameters, and every sequence $\xx$ ends with a distinguished end-of-sentence symbol $x_n$.

We implement character-level language models with Long Short-Term Memory (\textsc{lstm}) networks \citep{hochreiter1997long}. These encode the entire history $\xx_{<t}$ as a fixed-length vector $\vh_t$ by manipulating a memory cell $\cc_t$ through a set of gates. Then we define
\begin{equation} \label{eq:lstm}
p(x_t \mid \xx_{<t}, \ww) = \text{softmax}(\mathbf{W}\,\hh_t + \mathbf{b}).
\end{equation}
\textsc{lstm}s have an advantage over other recurrent architectures as memory gating mitigates the problem of vanishing gradients and captures long-distance dependencies \citep{pascanu2013difficulty}.

\section{Neural Language Modeling with a Universal Prior}
\label{sec:unversalprior}
The fundamental hypothesis of this work is that there exists a prior $p(\mathbf{w})$ over the weights of a neural language model
that places high probability on networks that describe human-like languages. Such a prior would provide an inductive bias that facilitates learning \emph{unseen} languages.
In practice, we construct it as the posterior distribution over the weights of a language model of \emph{seen} languages. Let $\calDtrain_\ell$ be the examples in language $\ell$, and let the examples in all training languages be $\calDtrain = \cup_{\ell \in \mathcal{T}} \calDtrain_\ell$. Taking a Bayesian approach, the posterior over weights is given by Bayes' rule: 
\begin{equation} \label{eq:bayesrule}
   \underbrace{p(\ww \mid \calDtrain) \vphantom{\prod_{\ell \in \calDtrain}}}_\textit{posterior} \propto \underbrace{\prod_{\ell \in \train} p(\calDtrain_\ell \mid \ww)}_{\textit{likelihood}} \, \underbrace{p(\ww) \vphantom{\prod_{\ell \in \calDtrain}}}_{\textit{prior}}
\end{equation}

\noindent
We take the prior of \cref{eq:bayesrule} to be a Gaussian with zero mean and covariance matrix $\sigma^2\,\textbf{I}$, i.e. 
\begin{equation} \label{eq:uninf_prior}
    p(\ww) = \frac{1}{\sqrt{2\pi\sigma^2}} \exp\left(-\frac{1}{2\sigma^2}||\ww||_2^2\right).
\end{equation}
However, computation of the posterior $p(\ww \mid \calDtrain)$ is woefully intractable: recall that, in our setting, each $p(\xx \mid \ww)$ is an LSTM language model, like the one defined in \cref{eq:lstm}. Hence, we opt for a simple approximation of the posterior, using the classic Laplace method \citep{mackay1992practical}. This method has recently been applied to other transfer learning or continuous learning scenarios in the neural network literature \cite{kirkpatrick2017overcoming,kochurov2018bayesian,ritter2018scalable}.

In \cref{ssec:laplace}, we first introduce the Laplace method, which approximates the posterior with a Gaussian centered at the maximum-likelihood estimate.\footnote{Note that, in general, the true posterior is multi-modal. The Laplace method instead approximates it with a unimodal distribution.} Its covariance matrix is amenable to be computed with backpropagation, as detailed in \cref{ssec:computation}. Finally, we describe how to use this distribution as a prior to perform maximum-a-posteriori inference over new data in \cref{ssec:MAP}.

\subsection{Laplace Method}
\label{ssec:laplace}
First, we (locally) maximize
the logarithm of the RHS of \cref{eq:bayesrule}:
\begin{equation}
\mathcal{L}(\ww) = \sum_{\ell \in \train} \log p(\calDtrain_{\ell} \mid \ww) + \log p(\ww)
\end{equation}
We note that $\mathcal{L}(\ww)$ is equivalent to the log-posterior up to an additive constant, i.e.\
\begin{equation}
    \log p(\ww \mid \calDtrain) = \mathcal{L}(\ww) - \log p(\calDtrain)
\end{equation}
where the constant $\log p(\calDtrain)$ is the log-normalizer.
Let $\ww^\star$ be a local maximizer of $\mathcal{L}$.\footnote{
In practice, non-convex optimization is only guaranteed to reach a critical point, which could be a saddle point. However, the derivation of Laplace's method assumes that we do reach a maximizer.}
We now approximate the log-posterior with a second-order Taylor expansion around $\ww^\star$:
\begin{align}\label{eq:log-posterior}
    &\log p(\ww \mid \calDtrain) \approx  \\
    \mathcal{L}(\ww^\star) + \frac{1}{2}(\ww - &\ww^\star)^{\top} \bH\,(\ww - \ww^\star) - \log p(\calDtrain) \nonumber
\end{align}
where $\mathbf{H}$ is the Hessian matrix. Note that we have omitted the first-order term, since the gradient $\nabla \mathcal{L}(\ww) = 0$ at the local maximizer $\ww^\star$.
This quadratic approximation to the log-posterior is Gaussian, which can be seen
by exponentiating the RHS of \cref{eq:log-posterior}:
\begin{align}
    & \frac{\exp\bigl[- \frac{1}{2} (\ww - \ww^\star)^\top (-\bH) (\ww - \ww^\star)\bigr]}{\sqrt{(2\pi)^{d} \left|- \bH\right|^{-1}}} \nonumber \\
\triangleq \, \, &\mathcal{N}(\ww^\star, -\bH^{-1})
\end{align}
where $\exp(\mathcal{L}(\ww^\star))$
is simplified from both numerator and denominator. Since $\ww^\star$ is a local maximizer, $\bH$ is a negative semi-definite matrix.\footnote{Note that, as a result, our representation of the Gaussian is non-standard; generally, the precision matrix is positive semi-definite.} The full derivation is given in \cref{app:laplace}.

In principle, computing the Hessian is possible by running backpropagation twice: This yields a matrix with $d^2$ entries. However, in practice, this is not possible. First, running backpropagation twice is tedious. Second, we can not easily store a matrix with $d^2$ entries since $d$ is the number of parameters in the language model, which is exceedingly large. 

\subsection{Approximating the Hessian}
\label{ssec:computation}
To cut the computation down to one pass, we exploit a property from theoretical statistics: Namely, that the Hessian of the log-likelihood bears a close resemblance
to a quantity known as the Fisher information matrix. This connection
allows us to develop a more efficient algorithm that approximates the Hessian with one pass of backpropagation.

We derive this approximation to the Hessian of $\calL(\ww)$ here. 
First, we note that due to the linearity of $\nabla^2$, we have
\begin{align} 
    \bH &= \nabla^2 \calL(\ww) \nonumber \\
    &= \nabla^2 \left(\sum_{\ell \in \train} \log p(\calDtrain_{\ell} \mid \ww) + \log p(\ww)\right) \nonumber \\
     &= \underbrace{\sum_{\ell \in \train}\nabla^2 \log p(\calDtrain_{\ell} \mid \ww)}_{\textit{likelihood}} + \underbrace{\vphantom{\sum_{\ell} \in \calDtrain} \nabla^2 \log p(\ww)}_\textit{prior} \label{eq:expfisher}
\end{align}
Note that the integral over languages $\ell \in \train$ is a discrete
summation, so we may exchange addends and derivatives such as is required for the proof. 

We now discuss each term of \cref{eq:expfisher} individually. 
First, to approximate the likelihood term, we draw on the relation between the Hessian and the Fisher information matrix.
A basic fact from information theory \cite{Cover2006} gives us that
the Fisher information matrix may be written in two equivalent 
ways: 
\begin{align}
    -&\mathop{{}\E} \left[\nabla^2 \log p(\calDtrain \mid \ww) \right] \\
    &=\underbrace{\mathop{{}\E} \left[ \nabla \log p(\calDtrain \mid \ww)\,\nabla \log p(\calDtrain \mid \ww)^{\top}  \right]}_{\textit{expected Fisher information matrix}} \nonumber
\end{align}
This equality suggests a natural approximation of the expected
Fisher information matrix---the \textit{observed} Fisher information matrix
\begin{align}
    -&\frac{1}{|\calDtrain|} \sum_{\xx \in \calDtrain} \nabla^2 \log p(\xx \mid \ww) \\
    &\approx \underbrace{ \frac{1}{|\calDtrain|} \sum_{\xx \in \calDtrain}  \nabla \log p(\xx \mid \ww)\,\nabla \log p(\xx \mid \ww)^{\top} }_{\textit{observed Fisher information matrix}} \nonumber
\end{align}
which is tight in the limit as $|\calDtrain| \rightarrow \infty$ due to the law of large numbers. Indeed, when
we have a large number of training exemplars,
the average of the outer products of the gradients 
will be a good approximation to the Hessian.
However, even 
this approximation still has $d^2$ entries, which is far too many to be practical. Thus, we further use a diagonal approximation. We denote the diagonal of the observed Fisher information matrix as the vector $\mathbf{f} \in \mathbb{R}^d$, which
we define as
\begin{align}
     \mathbf{f} =  \sum_{\ell \in \train} \sum_{\xx \in \calDtrain_\ell} \frac{1}{|\train| \cdot |\calDtrain_\ell|}  \bigl[\nabla \log p(\xx \mid \ww) \bigr]^2 \label{eq:hes_loglik}
\end{align}
where the $(\cdot)^2$ is applied point-wise.
Computation of the Hessian of the prior term in \cref{eq:expfisher} is more straightforward and does not require approximation. Indeed, generally, this is the negative inverse of the covariance matrix, which in our case means 
\begin{align}
\nabla^2 \log p(\ww) 
= - \frac{1}{\sigma^2} \mathbf{I} \label{eq:hes_logpri}
\end{align}
Summing the (approximate) Hessian of the log-likelihood in \cref{eq:hes_loglik} and the Hessian of the prior in \cref{eq:hes_logpri} yields our approximation to the Hessian of the log-posterior 
\begin{equation}
  \tilde{\bH} = -\textrm{diag}(\mathbf{f}) - \frac{1}{\sigma^2}\mathbf{I}
\end{equation}
The full derivation of the approximated Hessian is available in \cref{app:hessian}.

\subsection{MAP Inference}
\label{ssec:MAP}
Finally, we harness the posterior $p(\ww \mid \calDtrain) \approx \mathcal{N}(\ww^\star, - \tilde{\bH}^{-1})$ 
as the prior over model parameters for training a language model on new, held-out languages via MAP estimation. This is only an approximation to full Bayesian inference, because it does not characterize the entire distribution of the posterior, just the mode \citep{gelman2013bayesian}.

In the zero-shot setting, this boils down to using the mean of the prior $\ww^\star$ as network parameters during evaluation. In the few-shot setting, instead, we assume that some data for the target language $\ell \in \eval$ is available. Therefore, we maximize the log-likelihood given the target language data plus a regularizer that incarnates the prior, scaled by a factor of $\lambda$:
\begin{align} \label{eq:map}
    \mathcal{L}(\ww) = \sum_{\ell \in \eval} &\log p(\calDtrain_{\ell} \mid \ww) \\
    &+\frac{\lambda}{2} (\mathbf{w} - \mathbf{w}^\star)^{\top}\tilde{\bH}\,(\mathbf{w} - \mathbf{w}^\star) \nonumber
\end{align}
We denote the the prior $\mathcal{N}(\ww^\star, - \tilde{\bH}^{-1})$ 
that features in \cref{eq:map} as \textsc{Univ}, as it incorporates universal linguistic knowledge.
As a baseline for this objective, we perform MAP inference with an uninformative prior $\mathcal{N}(\textbf{0}, \textbf{I})$, which we label $\textsc{Ninf}$. In the zero-shot setting, this means that the parameters are sampled from the uninformative prior. In the few-shot setting, we maximize
\begin{align} \label{eq:ninf}
    \mathcal{L}(\ww) = \sum_{\ell \in \eval} &\log p(\calDtrain_{\ell} \mid \ww) - \frac{\lambda}{2} || \mathbf{w}||_2^2
\end{align} 
Note that, owing to this formulation, the uninformed $\textsc{Ninf}$ model does not have access to the posterior of the weights given the data from the training languages.

Moreover, as an additional baseline, we consider a common approach for transfer learning in neural networks \citep{ruder2017overview}, namely `fine-tuning.' After finding the maximum-likelihood value $\ww^\star$ on the training data, this is simply used to initialize the weights before further optimizing them on the held-out data. We label this method \textsc{FiTu}.

\section{Language Modeling Conditioned on Typological Features}
\label{sec:typcond}

Realistically, the prior over network weights should also be augmented with side information about the general properties of the held-out language to be learned, if such information is available. In fact, linguists have documented such information even for languages without plain digital texts available and stored it in the form of attribute--value features in publicly accessible databases \citep{croft2002typology,wals-2013}. 

The usage of such features to inform neural NLP models is still scarce, partly because the evidence in favor of their effectiveness is mixed \citep{Ponti:2018acl,ponti2018modeling}. In this work, we propose a way to distantly supervise the model with this \textit{side information} effectively. We extend our non-conditional language models outlined in \cref{sec:unversalprior} (\textsc{Bare}) to a series of variants \textit{conditioned} on language-specific properties, inspired by \citet{ostling2017continuous} and \citet{platanios2018contextual}. A fundamental difference from these previous works, however, is that they learn such properties in an end-to-end fashion from the data in a joint multilingual learning setting. Obviously, this is not feasible for the zero-shot setting and unreliable for the few-shot setting. Rather, we represent languages with their typological feature vector, which we assume to be readily available both for both training and held-out languages.

Let $\mathbf{t}_{\ell} \in [0, 1]^f$ be a vector of $f$ typological features for language $\ell \in \train \sqcup \eval$. We reinterpret the conditional language models within the Bayesian framework by estimating their posterior probability
\begin{equation}
    p(\ww \mid \calDtrain, \mathcal{F}) \propto \prod_{\ell \in \train} p(\calDtrain_\ell \mid \ww) \, p(\ww \mid \mathbf{t}_{\ell})
\end{equation} 
We now consider two possible methods to estimate $p(\ww \mid \mathbf{t}_{\ell})$. For both of them, we first encode the features through a non-linear transformation $f(\mathbf{t}_{\ell}) = \textrm{ReLU}(\mathbf{W}\,\mathbf{t}_{\ell} + \mathbf{b})$, where $\mathbf{W} \in \mathbb{R}^{r \times f}$ and $\mathbf{b} \in \mathbb{R}^{r}$, $r \ll f$. A first variant, labeled \textsc{Oest}, is based on \citet{ostling2017continuous}. Assuming the standard LSTM architecture where $\mathbf{o}_t$ is the output gate and $\cc_t$ is the memory cell, we modify the equation for the hidden state $\hh_t$ as follows:
\begin{equation}
    \hh_t = \bigl( {\bf o}_t \odot \tanh (\cc_t) \bigr) \oplus f(\mathbf{t}_{\ell})
\end{equation} 
where $\odot$ stands for the Hadamard product and $\oplus$ for concatenation. In other words, we concatenate the typological features to all the hidden states.

Moreover, we experiment with a second variant where the parameters of the LSTM are generated by a hyper-network (i.e., a simple linear layer with weight $\mathbf{W} \in \mathbb{R}^{|\ww| \times r}$) that transforms $f(\mathbf{t}_{\ell})$ into $\ww$. This approach, labeled \textsc{Plat}, is inspired by \citet{platanios2018contextual}, with the difference that they generate parameters for an encoder-decoder architecture for neural machine translation. 

On the other hand, we do not consider the conditional model proposed by \citet{sutskever2014sequence}, where $f(\mathbf{t}_\ell)$ would be used to initialize the values for $\hh_0$ and $\cc_0$. During the evaluation, for all time steps $t$, $\hh_t$ and $\cc_t$ are never reset on sentence boundaries, so this model would find itself at a disadvantage because it would require either to erase the sequential history cyclically or to lose memory of the typological features.

\section{Experimental Setup}
\label{sec:datamethod}
\paragraph{Data} The source for our textual data is the Bible corpus\footnote{\url{http://christos-c.com/bible/}} \citep{christodouloupoulos2015massively}.\footnote{This corpus is arguably representative of the variety of the world's languages: it covers 28 families, several geographic areas (16 languages from Africa, 23 from Americas, 26 from Asia, 33 from Europe, 1 from Oceania), and endangered or poorly documented languages (39 with less than 1M speakers).} We exclude languages that are not written in the Latin script and duplicate languages, resulting in a sample of 77 languages.\footnote{These are identified with their 3-letter \textsc{iso 639-3} codes throughout the paper. For the corresponding language names, consult \url{www.iso.org/standard/39534.html}.} Since not all translations cover the entire Bible, they vary in size. The text from each language is split into training, development, and evaluation sets (80-10-10 percent, respectively). Moreover, to perform MAP inference in the few-shot setting, we randomly sample 100 sentences from the train set of each held-out language.

We obtain the typological feature vectors from URIEL \citep{littell2017uriel}.\footnote{\url{www.cs.cmu.edu/~dmortens/uriel.html}} We include the features related to 3 levels of linguistic structure, for a total of 245 features: i) syntax, e.g. whether the subject tends to precede the object. These originate from the World Atlas of Language Structures \citep{wals-2013} and the Syntactic Structures of the World's Languages \citep{sswl}; ii) phonology, e.g.\ whether a language has distinctive tones; iii) phonological inventories, e.g.\ whether a language possesses the retroflex approximant \textipa{/\:R/}. Both ii) and iii) were originally collected in PHOIBLE \citep{phoible}. Missing values are inferred as a weighted average of the 10 nearest neighbor languages in terms of family, geography, and typology.

\begin{table*}[p]
\begin{center}
\def\arraystretch{0.83}
\begin{tabularx}{\linewidth}{>{\em}c rrr || >{\em}c rrr || >{\em}c rrr}
\toprule
& \textsc{\bfseries Ninf} & \multicolumn{2}{c||}{\textsc{ \bfseries Univ}} & & \textsc{\bfseries Ninf} & \multicolumn{2}{c||}{\textsc{ \bfseries Univ}} & & \textsc{\bfseries Ninf} & \multicolumn{2}{c}{\textsc{ \bfseries Univ}}  \\
 & \textsc{\bfseries Bare} & \textsc{\bfseries Bare} & \textsc{\bfseries Oest} & & \textsc{\bfseries Bare} & \textsc{\bfseries Bare} & \textsc{\bfseries Oest} & & \textsc{\bfseries Bare} & \textsc{\bfseries Bare} & \textsc{\bfseries Oest} \\
\hline
\cellcolor{blue!5}acu & 8.491 & \textbf{3.244} & 3.472 & \cellcolor{yellow!5}fra & 8.587 & \textbf{4.066} & 4.467 & \cellcolor{blue!5}por & 8.491 & \textbf{3.751} & 4.219\\
\cellcolor{red!5}afr & 8.607 & \textbf{3.229} & 3.995 & \cellcolor{green!5}gbi & 8.610 & \textbf{3.823} & 3.912 & \cellcolor{red!5}pot & 8.600 & \textbf{5.336} & 5.359\\
\cellcolor{red!5}agr & 8.603 & \textbf{3.779} & 3.946 & \cellcolor{blue!5}gla & 8.490 & 4.179 & 3.956 & \cellcolor{yellow!5}ppk & 8.596 & \textbf{4.506} & 4.599\\
\cellcolor{green!5}ake & 8.602 & \textbf{5.753} & 6.281 & \cellcolor{red!5}glv & 8.606 & \textbf{4.349} & 4.612 & \cellcolor{green!5}quc & 8.605 & \textbf{4.063} & 4.118\\
\cellcolor{blue!5}alb & 8.490 & \textbf{4.571} & 5.017 & \cellcolor{yellow!5}hat & 8.594 & \textbf{4.186} & 4.620 & \cellcolor{blue!5}quw & 8.488 & \textbf{3.560} & 4.027\\
\cellcolor{red!5}amu & 8.610 & \textbf{4.912} & 5.959 & \cellcolor{green!5}hrv & 8.606 & 4.050 & \textbf{3.441} & \cellcolor{red!5}rom & 8.603 & \textbf{3.669} & 4.056\\
\cellcolor{yellow!5}bsn & 8.591 & \textbf{5.046} & 5.695 & \cellcolor{blue!5}hun & 8.493 & \textbf{4.836} & 5.030 & \cellcolor{yellow!5}ron & 8.588 & \textbf{5.011} & 5.690\\
\cellcolor{green!5}cak & 8.603 & \textbf{4.068} & 4.326 & \cellcolor{red!5}ind & 8.604 & \textbf{3.796} & 4.311 & \cellcolor{green!5}shi & 8.601 & \textbf{5.496} & 5.946\\
\cellcolor{blue!5}ceb & 8.488 & \textbf{3.668} & 3.850 & \cellcolor{yellow!5}isl & 8.596 & \textbf{5.039} & 5.629 & \cellcolor{blue!5}slk & 8.491 & \textbf{4.304} & 4.512\\
\cellcolor{red!5}ces & 8.600 & \textbf{4.369} & 4.461 & \cellcolor{green!5}ita & 8.605 & 4.023 & \textbf{3.752} & \cellcolor{red!5}slv & 8.604 & \textbf{3.661} & 4.106\\
\cellcolor{yellow!5}cha & 8.594 & 4.366 & \textbf{4.353} & \cellcolor{blue!5}jak & 8.488 & \textbf{4.051} & 4.793 & \cellcolor{yellow!5}sna & 8.596 & \textbf{4.146} & 4.283\\
\cellcolor{green!5}chq & 8.598 & \textbf{6.940} & 7.623 & \cellcolor{red!5}jiv & 8.601 & \textbf{3.866} & 4.039 & \cellcolor{green!5}som & 8.614 & \textbf{4.159} & 4.470\\
\cellcolor{blue!5}cjp & 8.494 & \textbf{4.600} & 4.985 & \cellcolor{yellow!5}kab & 8.596 & \textbf{4.659} & 5.400 & \cellcolor{blue!5}spa & 8.489 & \textbf{3.645} & 4.020\\
\cellcolor{red!5}cni & 8.604 & \textbf{3.740} & 4.651 & \cellcolor{green!5}kbh & 8.607 & \textbf{4.663} & 4.950 & \cellcolor{red!5}srp & 8.604 & \textbf{3.414} & 3.437\\
\cellcolor{yellow!5}dan & 8.593 & \textbf{3.471} & 4.599 & \cellcolor{blue!5}kek & 8.491 & \textbf{4.666} & 4.944 & \cellcolor{yellow!5}ssw & 8.593 & 4.064 & \textbf{3.780}\\
\cellcolor{green!5}deu & 8.599 & \textbf{4.102} & 4.214 & \cellcolor{red!5}lat & 8.601 & \textbf{3.703} & 4.093 & \cellcolor{green!5}swe & 8.605 & 4.210 & \textbf{3.892}\\
\cellcolor{blue!5}dik & 8.490 & \textbf{4.447} & 4.533 & \cellcolor{yellow!5}lav & 8.588 & \textbf{5.415} & 6.130 & \cellcolor{blue!5}tgl & 8.487 & \textbf{3.639} & 3.878\\
\cellcolor{red!5}dje & 8.603 & \textbf{3.725} & 3.996 & \cellcolor{green!5}lit & 8.602 & \textbf{4.794} & 4.853 & \cellcolor{red!5}tmh & 8.602 & 4.830 & \textbf{4.711}\\
\cellcolor{yellow!5}djk & 8.592 & \textbf{3.663} & 3.874 & \cellcolor{blue!5}mam & 8.488 & \textbf{4.292} & 5.076 & \cellcolor{yellow!5}tur & 8.592 & \textbf{5.574} & 5.935\\
\cellcolor{green!5}dop & 8.609 & \textbf{5.950} & 7.351 & \cellcolor{red!5}mri & 8.606 & \textbf{3.440} & 4.074 & \cellcolor{green!5}usp & 8.604 & \textbf{4.127} & 4.337\\
\cellcolor{blue!5}eng & 8.488 & \textbf{3.816} & 4.028 & \cellcolor{yellow!5}nhg & 8.588 & \textbf{4.323} & 4.450 & \cellcolor{blue!5}vie & 8.490 & \textbf{7.137} & 7.484\\
\cellcolor{red!5}epo & 8.605 & \textbf{3.818} & 4.116 & \cellcolor{green!5}nld & 8.601 & \textbf{3.851} & 4.326 & \cellcolor{red!5}wal & 8.605 & \textbf{4.027} & 4.585\\
\cellcolor{yellow!5}est & 8.606 & \textbf{6.807} & 8.261 & \cellcolor{blue!5}nor & 8.492 & \textbf{3.174} & 3.902 & \cellcolor{green!5}wol & 8.607 & \textbf{4.290} & 4.420\\
\cellcolor{green!5}eus & 8.605 & \textbf{4.118} & 4.321 & \cellcolor{red!5}pck & 8.603 & \textbf{4.053} & 4.233 & \cellcolor{green!5}xho & 8.602 & \textbf{4.171} & 4.276\\
\cellcolor{blue!5}ewe & 8.490 & \textbf{5.049} & 5.497 & \cellcolor{yellow!5}plt & 8.603 & \textbf{4.364} & 4.648 & \cellcolor{blue!5}zul & 8.488 & \textbf{3.218} & 4.109\\
\cline{9-12}
\cellcolor{red!5}fin & 8.604 & \textbf{4.308} & 4.338 & \cellcolor{green!5}pol & 8.601 & \textbf{5.158} & 5.556 & \textsc{All}  & 8.572 & \textbf{4.343} & 4.691 \\

\bottomrule
\end{tabularx}
\end{center}
\vspace{-1.5mm}
\caption{BPC scores (lower is better) for the \textsc{Zero-shot} learning setting, with the uninformed prior (\textsc{Ninf}) and the universal prior (\textsc{Univ}): see \S\ref{sec:LSTM} for the descriptions of the priors. Note that for \textsc{Ninf} there is no difference between a \textsc{Bare} model and a conditional model (\textsc{Oest}). Colors define the partition in which each language (rows) has been held out.}
\label{tab:results_zeroshot}
\vspace{1cm}

\begin{center}
\def\arraystretch{0.83}
\begin{tabularx}{\linewidth}{>{\em}c r>{\bf}r || >{\em}c r>{\bf}r || >{\em}c rr || >{\em}c r>{\bf}r}
\toprule
 & \textsc{\bfseries Bare} & \textsc{Oest} & & \textsc{\bfseries Bare} & \textsc{Oest} & & \textsc{\bfseries Bare} & \textsc{\bfseries Oest} & & \textsc{\bfseries Bare} & \textsc{Oest} \\
\hline
acu & 1.413 & 1.308 & eng & 1.355 & 1.350 & kek & \textbf{1.131} & 1.133 & slk & 1.844 & 1.754\\
afr & 1.471 & 1.457 & epo & 1.471 & 1.450 & lat & 1.792 & \textbf{1.758} & slv & 1.848 & 1.793\\
agr & 1.701 & 1.581 & est & 0.333 & 0.150 & lav & 2.146 & \textbf{1.931} & sna & 1.489 & 1.457\\
ake & 1.453 & 1.377 & eus & 1.763 & 1.635 & lit & 1.895 & \textbf{1.833} & som & 1.477 & 1.468\\
alb & 1.590 & 1.552 & ewe & 2.084 & 1.944 & mam & 1.654 & \textbf{1.548} & spa & 1.559 & 1.525\\
amu & 1.402 & 1.340 & fin & 1.716 & 1.680 & mri & 1.342 & \textbf{1.330} & srp & 1.832 & 1.756\\
bsn & 1.232 & 1.172 & fra & 1.465 & 1.432 & nhg & 1.302 & \textbf{1.238} & ssw & 1.890 & 1.697\\
cak & 1.281 & 1.221 & gbi & 1.398 & 1.331 & nld & 1.621 & \textbf{1.601} & swe & 1.619 & 1.595\\
ceb & 1.193 & 1.185 & gla & 3.403 & 1.839 & nor & 1.623 & \textbf{1.590} & tgl & 1.221 & 1.210\\
ces & 1.872 & 1.795 & glv & 1.932 & 1.644 & pck & 1.731 & \textbf{1.711} & tmh & 2.786 & 2.301\\
cha & 1.934 & 1.790 & hat & 1.480 & 1.454 & plt & 1.296 & \textbf{1.286} & tur & 1.801 & 1.773\\
chq & 1.265 & 1.220 & hrv & 2.059 & 1.974 & pol & 1.743 & \textbf{1.698} & usp & 1.290 & 1.214\\
cjp & 1.706 & 1.565 & hun & 1.887 & 1.847 & por & 1.586 & \textbf{1.552} & vie & 1.648 & 1.637\\
cni & 1.348 & 1.290 & ind & 1.356 & 1.336 & pot & 2.484 & \textbf{2.144} & wal & 1.561 & 1.457\\
dan & 1.727 & 1.693 & isl & 1.845 & 1.808 & ppk & 1.538 & \textbf{1.439} & wol & 2.053 & 1.890\\
deu & 1.532 & 1.512 & ita & 1.615 & 1.583 & quc & 1.393 & \textbf{1.291} & xho & 1.680 & 1.634\\
dik & 1.979 & 1.835 & jak & 1.415 & 1.322 & quw & 1.498 & \textbf{1.418} & zul & 1.880 & 1.620\\
\cline{10-12}
dje & 1.570 & 1.550 & jiv & 1.705 & 1.572 & rom & 1.706 & \textbf{1.587} & \textsc{All} & 1.652 & \textbf{1.550} \\
djk & 1.515 & 1.435 & kab & 1.955 & 1.791 & ron & 1.572 & \textbf{1.537} &  &  & \\
dop & 1.810 & 1.676 & kbh & 1.436 & 1.371 & shi & 2.057 & \textbf{1.903} &  &  & \\
\bottomrule
\end{tabularx}
\end{center}
\caption{BPC results (lower is better) for the \textsc{Joint} learning setting, with the uninformed \textsc{Ninf} prior. These results constitute the expected ceiling performance for language transfer models.}
\label{tab:results_joint}
\end{table*}

\begin{table*}[ht!]
\begin{center}
\def\arraystretch{0.83}
\begin{tabularx}{0.83\linewidth}{>{\em}c rrrr || >{\em}c rrrr }
\toprule
& \textsc{\bfseries Ninf} & \textsc{\bfseries FiTu} & \multicolumn{2}{c||}{\textsc{ \bfseries Univ}} & & \textsc{\bfseries Ninf} & \textsc{\bfseries FiTu} & \multicolumn{2}{c}{\textsc{ \bfseries Univ}} \\
 & \textsc{\bfseries Bare} & \textsc{\bfseries Oest} & \textsc{\bfseries Bare} &  \textsc{\bfseries Oest} & & \textsc{\bfseries Bare} & \textsc{\bfseries Oest} & \textsc{\bfseries Bare} & \textsc{\bfseries Oest}  \\

\cellcolor{blue!5}acu& 4.203 & \textbf{2.117} & 2.551 & {2.136}  & \cellcolor{green!5}kbh& 4.644 & 2.362 & 2.434 & \textbf{2.288} \\ 
\cellcolor{red!5}afr& 4.423 & 3.620 & 3.042 & \textbf{2.773}  & \cellcolor{blue!5}kek& 4.613 & 2.809 & 3.015 & \textbf{2.714} \\ 
\cellcolor{red!5}agr& 4.268 & 3.282 & 3.403 & \textbf{2.457}  & \cellcolor{red!5}lat& 4.239 & 4.342 & 3.416 & \textbf{3.202} \\ 
\cellcolor{green!5}ake& 4.318 & 2.168 & 2.238 & \textbf{2.180}  & \cellcolor{yellow!5}lav& 4.765 & \textbf{2.867} & 3.842 & {2.917} \\ 
\cellcolor{blue!5}alb& 4.544 & 3.186 & 3.302 & \textbf{3.084}  & \cellcolor{green!5}lit& 4.769 & 3.752 & \textbf{3.592} & 3.668 \\ 
\cellcolor{red!5}amu& 4.486 & 2.820 & 3.948 & \textbf{2.080}  & \cellcolor{blue!5}mam& 4.525 & \textbf{2.274} & 2.873 & {2.363} \\ 
\cellcolor{yellow!5}bsn& 4.546 & 1.861 & 2.678 & \textbf{1.850}  & \cellcolor{red!5}mri& 3.795 & 3.482 & 3.010 & \textbf{2.459} \\ 
\cellcolor{green!5}cak& 4.426 & 1.994 & 2.053 & \textbf{1.956}  & \cellcolor{yellow!5}nhg& 4.373 & 2.004 & 2.480 & \textbf{1.965} \\ 
\cellcolor{blue!5}ceb& 4.084 & 2.562 & 2.595 & \textbf{2.470}  & \cellcolor{green!5}nld& 4.469 & 3.008 & 2.908 & \textbf{2.903} \\ 
\cellcolor{red!5}ces& 4.984 & 4.651 & 4.190 & \textbf{3.680}  & \cellcolor{blue!5}nor& 4.453 & 3.152 & \textbf{2.954} & 3.054 \\ 
\cellcolor{yellow!5}cha& 4.329 & 2.546 & 2.899 & \textbf{2.525}  & \cellcolor{red!5}pck& 4.246 & 4.011 & 3.532 & \textbf{3.030} \\ 
\cellcolor{green!5}chq& 4.941 & \textbf{1.948} & 2.078 & {1.963}  & \cellcolor{yellow!5}plt& 4.201 & 2.532 & 2.742 & \textbf{2.490} \\ 
\cellcolor{blue!5}cjp& 4.424 & \textbf{2.389} & 2.880 & {2.393}  & \cellcolor{green!5}pol& 4.853 & 3.852 & \textbf{3.620} & 3.788 \\ 
\cellcolor{red!5}cni& 4.185 & 2.797 & 3.018 & \textbf{1.982}  & \cellcolor{blue!5}por& 4.446 & 3.231& 3.198 & \textbf{3.098}\\ 
\cellcolor{yellow!5}dan& 4.719 & 3.211 & \textbf{3.127} & 3.180  & \cellcolor{red!5}pot& 4.299 & 3.773& 3.944 & \textbf{2.763}\\ 
\cellcolor{green!5}deu& 4.589 & 3.103 & 3.007 & \textbf{2.953}  & \cellcolor{yellow!5}ppk& 4.439 & \textbf{2.220} & 2.736 & {2.236}\\ 
\cellcolor{blue!5}dik& 4.380 & \textbf{2.640} & 3.020 & {2.667}  & \cellcolor{green!5}quc& 4.538 & 2.154& 2.242 & \textbf{2.108}\\ 
\cellcolor{red!5}dje& 4.382 & 3.815 & 3.398 & \textbf{2.898}  & \cellcolor{blue!5}quw& 4.223 & 2.196& 2.547 & \textbf{2.158}\\ 
\cellcolor{yellow!5}djk& 4.130 & \textbf{2.064} & 2.446 & {2.085}  & \cellcolor{red!5}rom& 4.378 & 3.121& 3.257 & \textbf{2.455}\\ 
\cellcolor{green!5}dop& 4.508 & 2.506 & 2.562 & \textbf{2.448}  & \cellcolor{yellow!5}ron& 4.579 & 3.273& 3.734 & \textbf{3.216}\\ 
\cellcolor{blue!5}eng& 4.436 & 2.808 & 2.913 & \textbf{2.719}  & \cellcolor{green!5}shi& 4.509 & \textbf{2.963} & 3.092 & {2.970}\\ 
\cellcolor{red!5}epo& 4.469 & 3.609 & 3.511 & \textbf{2.825}  & \cellcolor{blue!5}slk& 4.873 & 3.722& 3.812 & \textbf{3.631}\\ 
\cellcolor{yellow!5}est& 3.618 & \textbf{1.952} & 2.487 & {1.962}  & \cellcolor{red!5}slv& 4.633 & 4.630& 3.527 & \textbf{3.501}\\ 
\cellcolor{green!5}eus& 4.354 & 2.628 & 2.705 & \textbf{2.567}  & \cellcolor{yellow!5}sna& 4.455 & 2.910& 3.114 & \textbf{2.870}\\ 
\cellcolor{blue!5}ewe& 4.590 & 2.806 & 3.336 & \textbf{2.786}  & \cellcolor{green!5}som& 4.257 & 3.048& \textbf{2.908} & 2.934\\ 
\cellcolor{red!5}fin& 4.385 & 4.339 & 3.830 & \textbf{3.312}  & \cellcolor{blue!5}spa& 4.507 & 3.223& 3.149 & \textbf{3.090}\\ 
\cellcolor{yellow!5}fra& 4.551 & 3.086 & 3.276 & \textbf{2.981}  & \cellcolor{red!5}srp& 4.561 & 4.467& \textbf{3.367} & 3.380\\ 
\cellcolor{green!5}gbi& 4.250 & 2.138 & 2.170 & \textbf{2.054}  & \cellcolor{yellow!5}ssw& 4.370 & 2.611& 2.924 & \textbf{2.570}\\ 
\cellcolor{blue!5}gla& 4.159 & \textbf{2.377} & 2.835 & {2.395}  & \cellcolor{green!5}swe& 4.657 & 3.266& 3.184 & \textbf{3.177}\\ 
\cellcolor{red!5}glv& 4.346 & 3.523 & 3.702 & \textbf{2.644}  & \cellcolor{blue!5}tgl& 4.060 & 2.546& 2.592 & \textbf{2.436}\\ 
\cellcolor{yellow!5}hat& 4.468 & 2.929 & 3.048 & \textbf{2.849}  & \cellcolor{red!5}tmh& 4.618 & 4.087& 4.218 & \textbf{3.125}\\ 
\cellcolor{green!5}hrv& 4.615 & 3.845 & 3.608 & \textbf{3.588}  & \cellcolor{yellow!5}tur& 4.846 & \textbf{3.509} & 4.282 & {3.552}\\ 
\cellcolor{blue!5}hun& 4.806 & 3.589 & 3.709 & \textbf{3.522}  & \cellcolor{green!5}usp& 4.529 & 2.114& 2.189 & \textbf{2.073}\\ 
\cellcolor{red!5}ind& 4.377 & 3.317 & 3.258 & \textbf{2.420}  & \cellcolor{blue!5}vie& 5.185 & 3.018& 3.751 & \textbf{3.015}\\ 
\cellcolor{yellow!5}isl& 4.744 & 3.174 & 3.703 & \textbf{3.101}  & \cellcolor{red!5}wal& 4.398 & 2.986& 3.623 & \textbf{2.278}\\ 
\cellcolor{green!5}ita& 4.370 & 3.384 & 3.196 & \textbf{3.178}  & \cellcolor{green!5}wol& 4.621 & 2.898& 2.968 & \textbf{2.826}\\ 
\cellcolor{blue!5}jak& 4.532 & 2.113 & 2.650 & \textbf{2.126}  & \cellcolor{green!5}xho& 4.561 & 3.415& \textbf{3.208} & 3.289\\ 
\cellcolor{red!5}jiv& 4.338 & 3.413 & 3.475 & \textbf{2.504}  & \cellcolor{blue!5}zul& 4.564 & 2.625& 2.866 & \textbf{2.622}\\ 
\cline{6-10}
\cellcolor{yellow!5}kab& 4.649 & \textbf{2.783} & 3.574 & {2.800}  & \textsc{All}& 4.467 & 3.007 & 3.120 & \textbf{2.731}\\

\bottomrule
\end{tabularx}
\end{center}
\caption{BPC scores (lower is better) for the \textsc{Few-shot} learning setting, with \textsc{Ninf}, \textsc{FiTu}  and \textsc{Univ} priors. Colors define the partition in which each language (rows) has been held out.}
\label{tab:results_fewshot}
\end{table*}

\paragraph{Language Model}
We implement the LSTM following the best practices and choosing the hyper-parameter settings indicated by \citet{merityRegOpt,merityAnalysis}. Specifically, we optimize the neural weights with Adam \citep{kingma2014adam} and a non-monotonically decayed learning rate: its value is initialized as $10^{-4}$ and decreases by a factor of 10 every $1/3$rd of the total epochs. The maximum number of epochs amounts to 6 for training on $\calDtrain_\train$, with early stopping based on development set performance, and the maximum number of epochs is 25 for few-shot learning on $\calDtrain_{\ell \in \eval}$.

For each iteration, we sample a language proportionally to the amount of its data: $p(\ell) \propto |\calDtrain_\ell|$, in order not to exhaust examples from resource-lean languages in the early phase of training. Then, we sample without replacement from $\calDtrain_\ell$ a mini-batch of 128 sequences with a variable maximum sequence length.\footnote{This avoids creating insurmountable boundaries to back-propagation through time \citep{tallec2017unbiasing}.} This length is sampled from a distribution $m \sim \mathcal{N}(\mu=125, \sigma=5)$.\footnote{The learning rate is therefore scaled by $\frac{\nint{m}}{\mu} \cdot \frac{|\calDtrain_\train|}{|\train| \cdot |\calDtrain_\ell|}$, where $\nint{\cdot}$ is an operator that rounds to the closest integer.} Each epoch ends when all the data sequences have been sampled.

We apply several techniques of dropout for regularization, including variational dropout \citep{gal2016theoretically}, which applies an identical mask to all the time steps, with $p=0.1$ for character embeddings and intermediate hidden states and $p=0.4$ for the output hidden states. DropConnect \citep{wan2013regularization} is applied to the model parameters $\textbf{U}$ of the first hidden layer with $p=0.2$.

Following \newcite{merityRegOpt}, the underlying language model architecture consists of 3 hidden layers with 1,840 hidden units each. The dimensionality of the character embeddings is 400. We tie input and output embeddings following \newcite{merityAnalysis}. For conditional language models, the dimensionality of $f(\mathbf{t}_\ell)$ is set to 115 for the \textsc{oest} method based on concatenation \citep{ostling2017continuous}, and 4 (due to memory limitations) in the \textsc{plat} method based on hyper-networks \citep{platanios2018contextual}. For the regularizer in \cref{eq:map}, we perform grid search over the hyper-parameter $\lambda$: we finally select a value of $10^5$ for \textsc{Univ} and $10^{-5}$ for \textsc{Ninf}.

\paragraph{Regimes of Data Paucity}
We explore different regimes of data paucity for the held-out languages:

\noindent $\bullet$ \textsc{Zero-shot} transfer setting: we split the sample of 77 languages into 4 partitions. The languages in each subset are held out in turn, and we use their test set for evaluation.\footnote{Holding out each language individually would not increase the sample of training languages significantly, while inflating the number of experimental runs needed.} For each subset, we further randomly choose 5 languages whose development set is used for validation. The training set of the rest of the languages is used to estimate a prior over network parameters via the Laplace approximation. 

\noindent $\bullet$ \textsc{Few-shot} transfer setting: on top of the zero-shot setting, we use the prior to perform MAP inference over a small sample (100 sentences) from the training set of each held-out language.

\noindent $\bullet$ \textsc{Joint} multilingual setting:
the data includes the full training set for all 77 languages, including held-out languages.  This serves as a ceiling for the model performance in cross-lingual transfer.

\section{Results and Analysis}
\label{sec:results}
The results for our experiments are grouped in \cref{tab:results_zeroshot} for the \textsc{Zero-shot} regime, in \cref{tab:results_fewshot} for the \textsc{Few-shot} regime, and in \cref{tab:results_joint} for the \textsc{Joint} multilingual regime, which constitutes a ceiling to cross-lingual transfer performances. The scores represent Bits Per Character \cite[BPC;][]{graves2013generating}: this metric is simply defined as the negative log-likelihood of test data divided by $\ln 2$. We compare the results along the following dimensions: 

\paragraph{Informativeness of Prior} 
Our main result is that the \textsc{Univ} prior consistently outperforms the \textsc{Ninf} prior across the board and by a large margin in both \textsc{zero-shot} and \textsc{few-shot} settings. The scores of the na\"ivest baseline, \textsc{Zero-shot Ninf Bare}, are considerably worse than both \textsc{Zero-shot Univ} models: this suggests that the transfer of information on character sequences is meaningful. The lowest BPC reductions are observed for languages like Vietnamese (15.94\% error reduction) or Highland Chinantec (19.28\%) where character inventories differ the most from other languages. Moreover, the \textsc{Zero-shot Univ} models are on a par or better than even the \textsc{Few-shot Ninf} models. In other words, the most helpful supervision comes from a universal prior rather than from a small in-language sample of sentences. This demonstrates that the \textsc{Univ} prior is truly imbued with universal linguistic knowledge that facilitates learning of previously unseen languages.
 
The averaged BPC score for the other baseline without a prior, \textsc{Fine-tune}, is 3.007 for \textsc{Few-shot Oest}, to be compared with 2.731 BPC of \textsc{Univ}. Note that fine-tuning is an extremely competitive baseline, as it lies at the core of most state-of-the-art NLP models \citep{peters2019tune}. Hence, this result demonstrates the usefulness of Bayesian inference in transfer learning.

\paragraph{Conditioning on Typological Information} Another important result regards the fact that conditioning language models on typological features yields opposite effects in the \textsc{zero-shot} and \textsc{few-shot} settings. Comparing the columns of the \textsc{Bare} and \textsc{Oest} models in \cref{tab:results_zeroshot} reveals that the non-conditional baseline \textsc{Bare} is superior for 71 / 77 languages (the exceptions being Chamorro, Croatian, Italian, Swazi, Swedish, and Tuareg). On the other hand, the same columns in \cref{tab:results_fewshot} and \cref{tab:results_joint} reveal an opposite pattern: \textsc{Oest} outperforms the \textsc{Bare} baseline in 70 / 77 languages. Finally, \textsc{Oest} surpasses the \textsc{Bare} baseline in the \textsc{Joint} setting for 76 / 77 languages (save Q'eqchi').

We also also take into consideration an alternative conditioning method, namely \textsc{Plat}. For clarity's sake, we exclude this batch of results from \cref{tab:results_zeroshot} and \cref{tab:results_fewshot}, as this method proves to be consistently worse than \textsc{Oest}. In fact, the average BPC of \textsc{Plat} amounts to 5.479 in the \textsc{Zero-shot} setting and 3.251 in the \textsc{Few-shot} setting. These scores have to be compared with 4.691 and 2.731 for \textsc{Oest}, respectively.

The possible explanation behind the mixed evidence on the success of typological features points to some intrinsic flaws of typological databases. \citet{ponti2018modeling} has shown how their feature granularity may be too coarse to liaise with data-driven probabilistic models, and inferring missing values due to the limited coverage of features results in additional noise. As a result, language models seem to be damaged by typological features in absence of data, whereas they benefit from their guidance when at least a small sample of sentences is available in the \textsc{Few-shot} setting. 

\paragraph{Data Paucity} Different regimes of data paucity display uneven levels of performance. The best models for each setting (\textsc{Zero-shot Univ Bare}, \textsc{Few-shot Univ Oest}, and \textsc{Joint Oest}) reveal large gaps between their average scores. Hence, in-language supervision remains the best option when available: transferred language models always lag behind their supervised equivalents.

\section{Related Work}
\label{sec:relatedwork}
LSTMs have been probed for their inductive bias towards syntactic dependencies \citep{linzen-etal-2016-assessing} and grammaticality judgments \citep{D18-1151,warstadt2018neural}. \citet{ravfogel2019studying} have extended the scope of this analysis to typologically different languages through \textit{synthetic} variations of English. In this work, we aim to model the inductive bias explicitly by constructing a prior over the space of neural network parameters.

Few-shot word-level language modeling for truly under-resourced languages such as Yongning Na has been investigated by \citet{adams2017cross} with the aid of a bilingual lexicon. \citet{vinyals2016matching} and \citet{munkhdalai2018metalearning} proposed novel architectures (Matching Networks and LSTMs augmented with Hebbian Fast Weights, respectively) for rapid associative learning in English, and evaluated them in few-shot cloze tests. In this respect, our work is novel in pushing the problem to its most complex formulation, zero-shot inference, and in taking into account the largest sample of languages for language modeling to date.

In addition to those considered in our work, there are also alternative methods to condition language models on features. \citet{kalchbrenner2013recurrent} used encoded features as additional biases in recurrent layers. \citet{kiros2014multimodal} put forth a log-bilinear model that allows for a `multiplicative interaction' between hidden representations and input features (such as images). With a similar device, but a different gating method, \citet{tsvetkov2016polyglot} trained a phoneme-level joint multilingual model of words conditioned on typological features from \citet{phoible}.

The use of the Laplace method for neural transfer learning has been proposed by \citet{kirkpatrick2017overcoming}, inspired by synaptic consolidation in neuroscience, with the aim to avoid catastrophic forgetting. \citet{kochurov2018bayesian} tackled the problem of continuous learning by approximating the posterior probabilities through stochastic variational inference. \citet{ritter2018scalable} substitute diagonal Laplace approximation with a Kronecker factored method, leading to better uncertainty estimates. Finally, the regularizer proposed by \citet{duong2015low} for cross-lingual dependency parsing can be interpreted as a prior for MAP estimation where the covariance is an identity matrix.

\section{Conclusions}
\label{sec:conclusions}
In this work, we proposed a Bayesian approach to transfer language models cross-lingually. We created a universal prior over neural network weights that is capable of generalizing well to new languages suffering from data paucity. The prior was constructed as the posterior of the weights given the data from available training languages, inferred via the Laplace method. Based on the results of character-level language modeling on a sample of 77 languages, we demonstrated the superiority of this prior imbued with universal linguistic knowledge over uninformative priors and unnormalizable priors (i.e., the widespread fine-tuning approach) in both zero-shot and few-shot settings. Moreover, we showed that adding language-specific side information drawn from typological databases to the universal prior further increases the levels of performance in the few-shot regime. While cross-lingual transfer still lags behind supervised learning when sufficient in-language data are available, our work is a step towards bridging this gap in the future.

\section*{Acknowledgements}
This work is supported by the ERC Consolidator Grant LEXICAL (no 648909). RR was partially funded by ISF personal grants No.\ 1625/18. 

\bibliography{tacl2018}
\bibliographystyle{acl_natbib}

\clearpage
\appendix
\section{Character Distribution} 
\begin{figure*}[t]
    \centering
    \includegraphics[width=\textwidth]{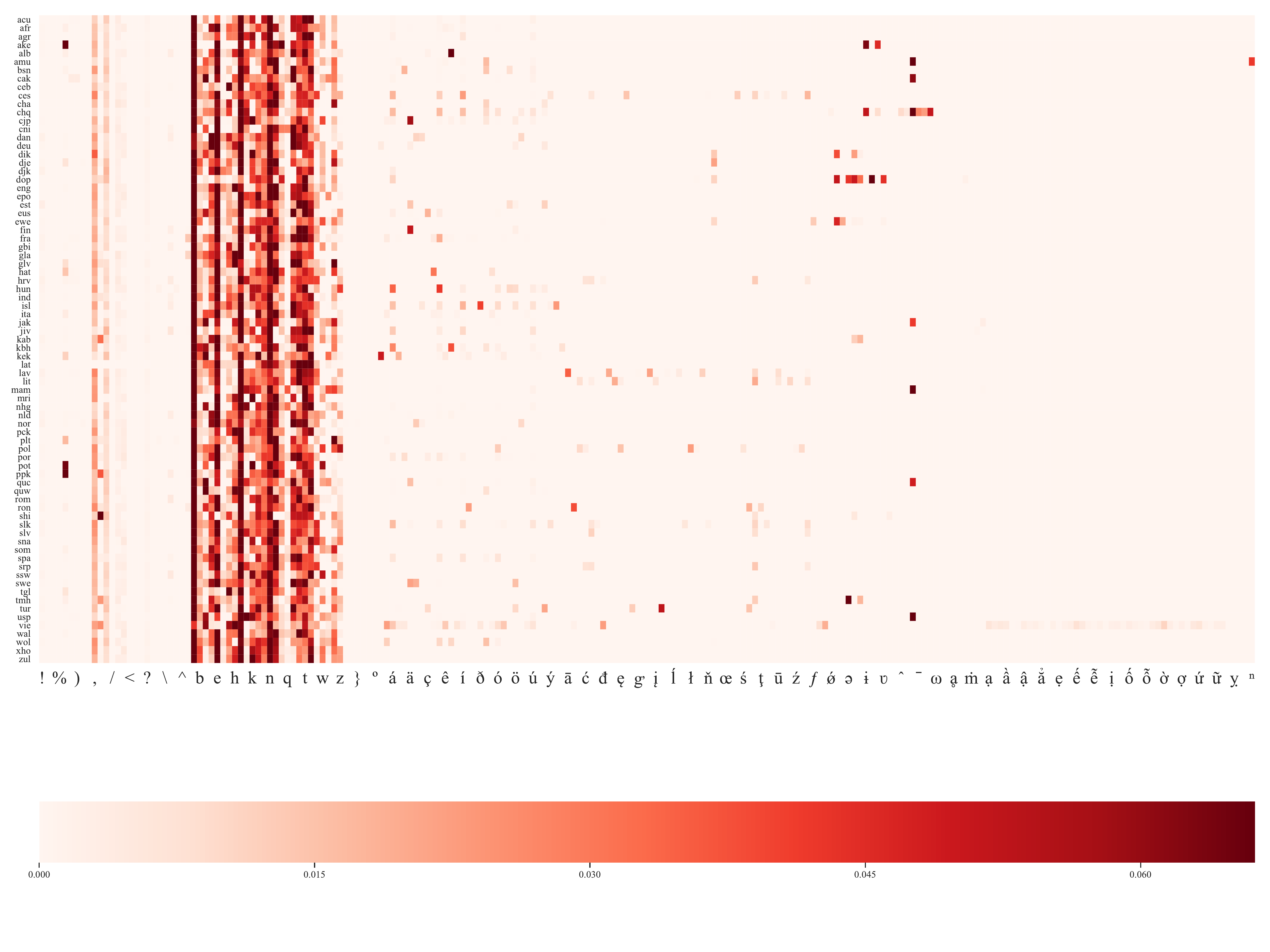}
    \vspace{-1.5cm}
    \caption{Unigram character distribution (x-axis) per language (y-axis). Note how some rows stand out as outliers.}
    \label{fig:heatchar}
\end{figure*}

Even within the same setting, BPC scores vary enormously across languages in both the \textsc{Zero-shot} and \textsc{Few-shot} settings, which requires an explanation. Similarly to \citet{gerz2018language,gerz2018relation}, we run a correlation analysis between language modeling performance and basic statistics of the data. In particular, we first create a vector of unigram character counts for each language, shown in \cref{fig:heatchar}. Then we estimate the cosine distance between the vector of each language and the average of all the others in our sample. This cosine distance is a measure of the `exoticness' of a language's character distribution.

Pearson's correlation between such cosine distance and the perplexity of \textsc{Univ Bare} in each language reveals a strong correlation coefficient $\rho = 0.53$ and a statistical significance of $p<10^{-6}$ in the \textsc{Zero-shot} setting. On the other hand, such correlation is absent ($\rho = -0.13$) and insignificant $p>0.2$ in the \textsc{Few-shot} setting. In other words, if a few examples of character sequences are provided for a target language, language modeling performance ceases to depend on its unigram character distribution.

\begin{figure*}[ht]
    \centering
    \includegraphics[width=\textwidth]{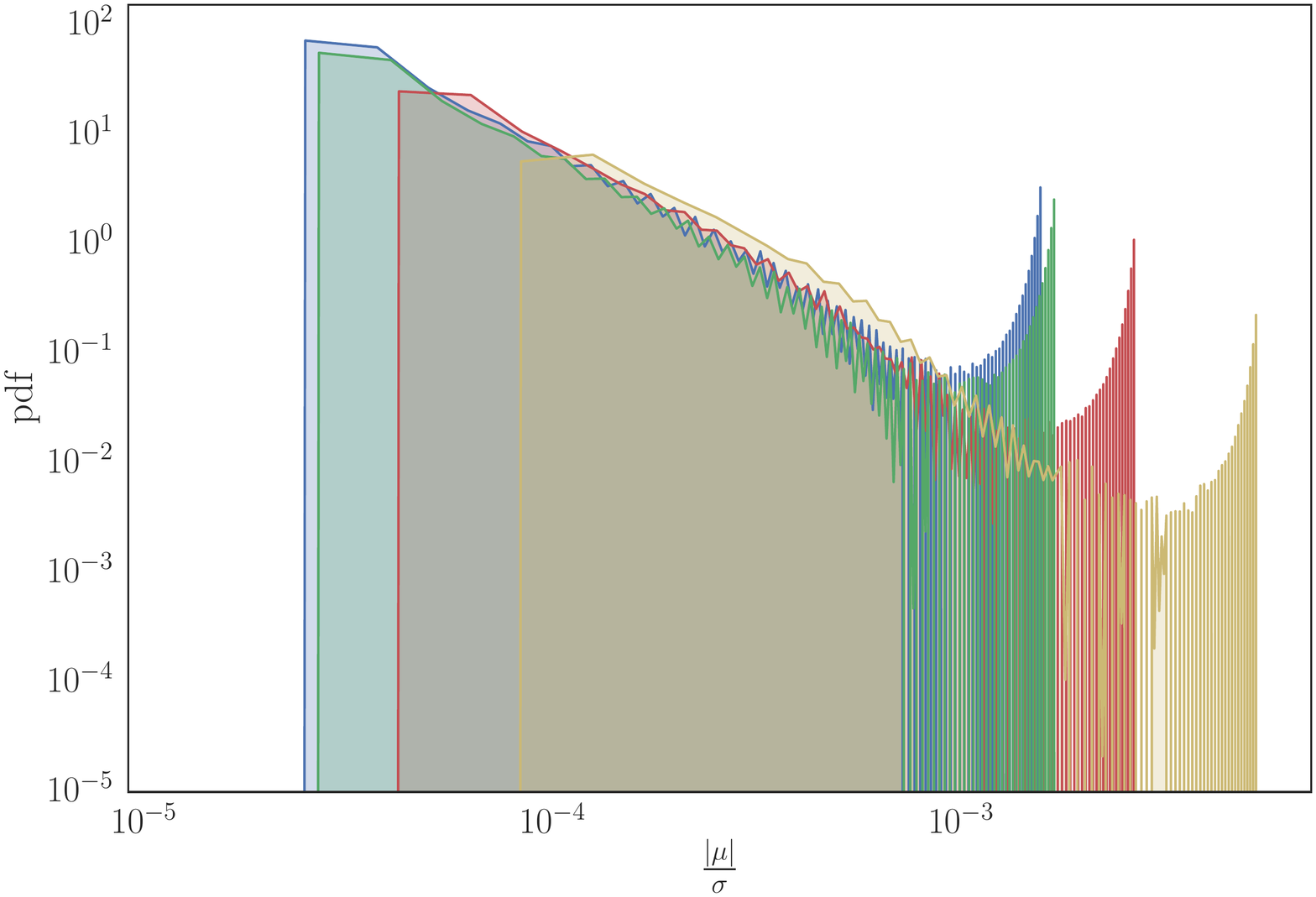}
    \caption[Signal-to-noise ratio of the learned posteriors.]{Probability density function of the signal-to-noise ratio for each parameter of the learned posteriors in the \textsc{Univ Bare} language models on splits 1 (blue), 2 (red), 3 (green), 4 (gold). The plot is in log-log scale. }
    \label{fig:stnposterior}
\end{figure*}

\section{Probing of Learned Posteriors}
Finally, it remains to establish which sort of knowledge is embedded in the universal prior. How to probe a probability distribution over weights in the non-conditional \textsc{Univ Bare} language model? First, we study the signal-to-noise ratio of each parameter $\ww_i$, computed as $\frac{|\mu_i|}{\sigma_i}$, in each of the 4 splits. Intuitively, this metric quantifies the `informativeness' of each parameter, which is proportional to both the absolute value of the mean and the inverse standard deviation of the estimate. The probability density function of the signal-to-noise ratio is shown in \cref{fig:stnposterior}. From this plot, it emerges that the estimated uncertainty is generally low (small $\sigma_i$ denominators yield high values). Most crucially, the signal-to-noise values concentrate on the left of the spectrum. This means that most weights will not incur any penalty for changing during few-shot learning based on \cref{eq:map}; on the other hand, there is a bulk of highly informative parameters on the right of the spectrum that are very likely to remain fixed, thus preventing catastrophic forgetting. All splits display such a pattern, although somewhat shifted.

\newcolumntype{Y}{>{\centering\arraybackslash}X}
\begin{table*}[p]
    \centering
    {
    \begin{tabularx}{\textwidth}{>{\sc}c|>{\em}Y|>{\sc}c|>{\em}Y}
lit &  javen šuksyr sun siriai tes pije nuks & shi & ereswrin an da$\gamma$tartnaas ni mad  yanó \\
nor & s hech far binje alrn bre a ver e hior  & jak & fi pelo ayok musam nejaz jih tewat ushi \\
kek & sx er taj chan linam laj âtebke naque  & swe & ssiar řades perdeshen heklui tart si a \\
jiv & da tum suuam s$\iota$tas nekkin una tekaru ni & dik & e w\textipa{E}n ke nu\textipa{\ng} ni piyitia de run ye e ke \\
dje & a ciya toi milkak mo to yen nga  suci & ewe & å mula pe ose le ake  mente amesa ke kul \\
slk & o je to temokoé lostave sa jesé gukli & alb & I kur je ki thet je ji tin nuk t tho  \\
ces & e je jek jem neute\textipa{\ng} rekssýj jazá níb ws & cni & u pen mireshisinoe airitcsa ateani yi \\
por &   uč somo ai jegparase saves e iper to & pot & neta ynimka nekin linaayi meu carií a \\
spa & esquár y lues dusme allis nencec adi & zul & \foreignlanguage{vietnamese}{ở}nakan kuná bencro krileke konusti k \\
glv &  ayr shżi ayn ai sephson a gil or geee  & quw & ai chimira kachisinyra poi apre asyu \\
pol & eteni na hidi c\foreignlanguage{vietnamese}{ế}ho oż swchj jeci i cil & agr & ji ica ama kujaa muri wajetar aumam hu \\
quc & ûs xe cä wija ro pio kin cbi' ij jejac & dop & bt\textipa{ElO} $\iota$ tela$\gamma$a \textipa{kO} n$\epsilon\iota$ zû$\gamma\iota$ \textipa{nEk@} \textipa{pO} \\
wal &  banjake la dos que benthi shivegina  & eus & cerer nagcermac istirinun qatserite \\
xho & ukayla azigeecoa kosubentisiili jen maky & hun &  elyet a bukot aky azraá ot mu háláj y \\
som & ao kun adku i sir jija i befey yadui  & gla & o e kere hhó sho dhöìr te ilailui a tu a \\
tgl &  ikugy peo asha atan kao amai kain ak a & pck & u gihiha ki mi dhia mea la hen a puh ih \\
cjp & pae yei aje kin trheka pän awawa ri s & afr & mal hoor in e sheei wer var buerkeas en \\
acu & animmhi mustatur tukaw aants aastasai a & usp &  okan mi ykis ris rajajkujij taka ja \\
fin & i koin suu meit ja  ii  soi tetot jasw & ind & t berka duhah menkad kemia ukus keri ya \\
mri & oki ka benoka ai ki kimanka pikaka ko & rom & hal kus seke nukertia dehe neshes hos n \\
slv & čičvim koko si neče pau ku meta noj ne  & tmh & \textipa{@}rofm sibarn awigtir $\epsilon$li d usi  leped \\
hrv &  ca ka te zet jon jem nezin  isak ve u  & ita & tri cordia io si si conse de namni nel \\
epo & j li inij keris ec xom el e sepon kaj  & srp & e se a nil do zasom kuz je sefe nij hoč  \\
amu & \'mibinya na ñero melee cano' ndo' cy'oc & nld &  e suet en de semeshord ak abaido zin  \\
kbh & \"{x}e aquangmomnaynangmuacha tojam  & lat & ifte quissi fetam remnas emens in timnex \\
ceb &  abithon kayay isa atoug giraban sula  & mam & í la \textipa{\ng}il a cheh tjea nut tej quxen kaj  \\
gbi & fuma ome pani de imoako kema kaye ntul & vie &  \foreignlanguage{vietnamese}{hẩ kì đãi bi  ầt ni $\gamma$ì sa hiổ vū r} \\
eng & g ban urse auth ahen ant msesher at nhe & & \\
\hline
isl & j noka nie leli maken ti aide ni itsim a & est & inam acha dius dempegun geben parug j \\
sna & xe yare ske tengker ci bendar nu derbe & cha & \^{e} duka ka kina kia nextis ne aka nisa  \\
ron &  ma awa nasil ko khe ni koy koj tikis t & fra & dis assan in man usia issokoj mulel e me \\
kab & je cana ka casa chomdis mear de ber h & djk &  okrana anginar matom iliantarinta a non \\
nhg & chun neyal den ma kashtaka asa as riste & lav & ilu kagsa eriri isi paj ewri bus os \\
dan & dnepse aa aye sas ningli inas giksaj abe & bsn &  as juhma yainawa nusa wali apai basti \\
ppk & ios yena mona kemewascoj ni ne maa & hat & a kuneati ua veskos oramaj meseqen ye k \\
ssw & nta yoti gesi kela nii ikasgaber ni tus & tur & che a shachmo \foreignlanguage{vietnamese}{ề}spi meng rinnaj e ish em \\
wol &  alen kokpan fed man benu pei ei kestam & ake & n jes silem semmo caja arka wagtoa doo  \\
deu &  ke giko si obi rer nin eber tun ke ele & chq & shas nej neysakun kina alistad mesabe  \\
cak &  tej je awem titoj lunik c'u chis m ni & plt & \textipa{V}wi meyak me imai anet alavis edte kin
    \end{tabularx}
    \caption[Examples of text generated from the learned language models.]{Randomly generated text on observed languages (top) and held-out languages (bottom) in the 4th split.}
    \label{tab:randngen}
    }
\end{table*}

Second, to study the effect of conditioning the universal prior on typological features, I generate random sequences of 25 characters from the learned prior in each language. The first character is chosen uniformly at random, and the subsequent ones are sampled from the distribution given by \cref{eq:nextcharprob} with a temperature of 1. The resulting texts are shown in \cref{tab:randngen}. Although this would warrant a more thorough and systematic analysis, from a cursory view it is evident of the sequences abide with universal phonological patterns, e.g.\ favoring vowels as syllabic nuclei and ordering consonants based on sonority hierarchy. Moreover, the language-specific information clearly steers predicted sequences towards the correct inventory of characters, as demonstrated by Vietnamese (\textsc{vie}) and Lukpa (\textsc{dop}) in \cref{tab:randngen}.

\clearpage
\onecolumn
\section{Derivation of the Laplace Approximation}
\label{app:laplace}
\begin{equation}
\begin{aligned}
p(\ww \mid \calDtrain) &= \frac{\exp\bigl(\mathcal{L}(\ww)\bigr)}{\int \exp\bigl(\mathcal{L}(\ww)\bigr) \, \mathrm{d}\ww} \quad \textit{Bayes rule}\\
&\approx \frac{\exp\bigl[\mathcal{L}(\ww^\star) + (\ww - \ww^\star)^\top \nabla\mathcal{L}(\ww^\star) + \frac{1}{2} (\ww - \ww^\star)^\top \bH \, (\ww - \ww^\star)\bigr]}{\int \exp\bigl[\mathcal{L}(\ww^\star) + (\ww - \ww^\star)^\top \nabla\mathcal{L}(\ww^\star) + \frac{1}{2} (\ww - \ww^\star)^\top \bH \, (\ww - \ww^\star)\bigr] \, \mathrm{d}\ww} \quad \textit{Taylor expansion} \\
&= \frac{\exp\bigl[\mathcal{L}(\ww^\star) + \frac{1}{2} (\ww - \ww^\star)^\top \bH \, (\ww - \ww^\star)\bigr]}{\int \exp\bigl[\mathcal{L}(\ww^\star) + \frac{1}{2} (\ww - \ww^\star)^\top \bH \, (\ww - \ww^\star)\bigr] \, \mathrm{d}\ww} \qquad \nabla\mathcal{L}(\ww)\rvert_{\ww^\star} = \mathbf{0} \\
&= \frac{\exp \bigl(\mathcal{L}(\ww^\star)\bigr) \exp\bigl[- \frac{1}{2} (\ww - \ww^\star)^\top (-\bH) \, (\ww - \ww^\star)\bigr]}{\exp \bigl(\mathcal{L}(\ww^\star)\bigr) \int \exp\bigl[- \frac{1}{2} (\ww - \ww^\star)^\top (- \bH) (\ww - \ww^\star)\bigr] \, \mathrm{d}\ww} \quad \textit{exponential of sum} \\
&= \frac{\exp\bigl[- \frac{1}{2} (\ww - \ww^\star)^\top (-\bH) (\ww - \ww^\star)\bigr]}{\sqrt{(2\pi)^{d} \left|- \bH\right|^{-1}}} \quad \textit{integration and simplification}\\
&\triangleq \mathcal{N}(\ww^\star, -\bH^{-1}) \\
\end{aligned}
\end{equation}

\section{Derivation of the Approximated Hessian}
\label{app:hessian}
We assume $\ww \sim \mathcal{N}(\mathbf{0}, \sigma^2\mathbf{I})$. Given the relationship among the expected Fisher Information $\mathcal{I}(\ww)$, the observed Fisher Information $\mathcal{J}(\ww)$, the observed Fisher Information based on $|\calDtrain|$ samples $\mathcal{J}_\calDtrain(\ww)$, and the Hessian $\bH$:
\begin{equation}
    - \mathcal{I}(\ww) = - \E\mathcal{J}(\ww) \approx - \frac{1}{|\calDtrain|} \mathcal{J}_{\calDtrain}(\ww) = 
    \frac{1}{|\calDtrain|} \, \bH = \frac{1}{|\calDtrain|} \nabla^2 \calL(\ww)
\end{equation}
we can derive our approximation of $\frac{1}{|\calDtrain|} \, \bH$:

\clearpage
{
\begin{equation}
\begin{aligned}
    & \quad \, \, \frac{1}{|\calDtrain|} \nabla^2 \calL(\ww) \\
    &= \frac{1}{|\calDtrain|} \nabla^2 \left(\sum_{\ell \in \train} \log p(\calDtrain_{\ell} \mid \ww) + \log p(\ww)\right) \quad \textit{definition of } \calL(\ww) \\
     &= \sum_{\ell \in \train} \sum_{\xx \in \calDtrain_\ell} \frac{1}{|\train| \cdot |\calDtrain_\ell|}  \nabla^2 \log p(\xx \mid \ww) + \nabla^2 \log p(\ww) \quad \textit{linearity of } \nabla^2  \\
     &= \sum_{\ell \in \train} \sum_{\xx \in \calDtrain_\ell} \frac{1}{|\train| \cdot |\calDtrain_\ell|} \nabla \left( \frac{\nabla p(\xx \mid \ww)}{ p(\xx \mid \ww)}\right) + \nabla^2 \log p(\ww) \quad \textit{derivative of logarithm} \\
     &= \sum_{\ell \in \train} \sum_{\xx \in \calDtrain_\ell} \frac{1}{|\train| \cdot |\calDtrain_\ell|} \frac{p(\xx \mid \ww) \nabla^2 p(\xx \mid \ww) - \nabla p(\xx \mid \ww) \nabla p(\xx \mid \ww)^\top}{p(\xx \mid \ww)^2} \\
     &\qquad + \nabla^2 \log p(\ww) \quad \textit{quotient rule} \\
     &= \sum_{\ell \in \train} \sum_{\xx \in \calDtrain_\ell} \frac{1}{|\train| \cdot |\calDtrain_\ell|}  \left[\frac{\nabla^2 p(\xx \mid \ww)}{p(\xx \mid \ww)} - \left(\frac{ \nabla p(\xx \mid \ww)}{p(\xx \mid \ww)}\right) \left(\frac{ \nabla p(\xx \mid \ww)}{p(\xx \mid \ww)}\right)^\top\right] \\
     &\qquad + \nabla^2 \log p(\ww) \quad \textit{rearrange and simplify} \\
     &= \sum_{\ell \in \train} \sum_{\xx \in \calDtrain_\ell} \frac{1}{|\train| \cdot |\calDtrain_\ell|} \left[\frac{\nabla^2 p(\xx \mid \ww)}{p(\xx \mid \ww)} - \nabla \log p(\xx \mid \ww) \, \nabla \log p(\xx \mid \ww)^\top\right] \\
     &\qquad + \nabla^2 \log p(\ww) \quad \textit{derivative of logarithm} \\
     &\approx \sum_{\ell \in \train} \frac{1}{|\train|} \left[ \E_{\xx \sim \, p(\cdot \mid \ww)}  \frac{\nabla^2 p(\xx \mid \ww)}{p(\xx \mid \ww)} - \frac{1}{|\calDtrain_\ell|} \sum_{\xx \in \calDtrain_\ell} \nabla \log p(\xx \mid \ww) \, \nabla \log p(\xx \mid \ww)^\top\right] \\
     &\qquad + \nabla^2 \log p(\ww) \quad \textit{sample average as expectation} \\
     &= \sum_{\ell \in \train} \frac{1}{|\train|} \left[ \int \frac{\nabla^2 p(\xx \mid \ww)}{p(\xx \mid \ww)} \, p(\xx \mid \ww) \, \mathrm{d}\xx - \frac{1}{|\calDtrain_\ell|} \sum_{\xx \in \calDtrain_\ell} \nabla \log p(\xx \mid \ww) \, \nabla \log p(\xx \mid \ww)^\top\right] \\
     &\qquad + \nabla^2 \log p(\ww) \quad \textit{expectation as integral} \\
     &= \sum_{\ell \in \train} \frac{1}{|\train|} \left[ \nabla^2 \int p(\xx \mid \ww) \, \mathrm{d}\xx - \frac{1}{|\calDtrain_\ell|} \sum_{\xx \in \calDtrain_\ell} \nabla \log p(\xx \mid \ww) \, \nabla \log p(\xx \mid \ww)^\top \right] \\
     &\qquad + \nabla^2 \log p(\ww) \quad \textit{simplify} \\
     &= \sum_{\ell \in \train} \sum_{\xx \in \calDtrain_\ell} \frac{- 1}{|\train| \cdot |\calDtrain_\ell|} \nabla \log p(\xx \mid \ww) \, \nabla \log p(\xx \mid \ww)^\top + \nabla^2 \log p(\ww) \quad \textit{derivative of constant} \\
     &\approx \sum_{\ell \in \train} \sum_{\xx \in \calDtrain_\ell} \frac{- 1}{|\train| \cdot |\calDtrain_\ell|} \textrm{diag}\biggl[\nabla \log p(\xx \mid \ww)\biggr]^2 + \nabla^2 \log p(\ww) \quad \textit{diagonal approximation} \\
     &= \sum_{\ell \in \train} \sum_{\xx \in \calDtrain_\ell} \frac{- 1}{|\train| \cdot |\calDtrain_\ell|} \textrm{diag}\biggl[\nabla \log p(\xx \mid \ww)\biggr]^2 - \frac{1}{\sigma^2}\mathbf{I} \quad \textit{second derivative of log-probability}
\end{aligned}
\end{equation}
}

\end{document}